%% file: main.tex
\documentclass[final]{cvpr}

\usepackage{times}
\usepackage{epsfig}
\usepackage{graphicx}
\usepackage{amsmath}
\usepackage{amssymb}

\usepackage[pagebackref=true,breaklinks=true,colorlinks,bookmarks=false]{hyperref}

\hypersetup{
     colorlinks=true,
     linkcolor=green,
     filecolor=blue,
     citecolor = blue,      
     urlcolor=cyan,
     }

\usepackage{balance}
\usepackage{amsmath,amssymb} 
\usepackage{color}
\usepackage{multicol}

\usepackage{bbm}
\usepackage{caption}
\usepackage{algorithm}
\usepackage[noend]{algpseudocode} 
\usepackage{bbding}
\usepackage[noadjust]{cite}
\usepackage{enumitem}

\usepackage{array}
\newcolumntype{L}[1]{>{\raggedright\let\newline\\\arraybackslash\hspace{0pt}}m{#1}}
\newcolumntype{C}[1]{>{\centering\let\newline\\\arraybackslash\hspace{0pt}}m{#1}}
\newcolumntype{R}[1]{>{\raggedleft\let\newline\\\arraybackslash\hspace{0pt}}m{#1}}

\def\cvprPaperID{6362} 
\def\confYear{CVPR 2021}

\begin{document}

\title{Revisiting Knowledge Distillation for Object Detection}


\author{Amin Banitalebi-Dehkordi\\
Huawei Technologies Canada Co., Ltd.\\
{\tt\small amin.banitalebi@huawei.com}
}

\maketitle

\input{00_abstract.tex}


\input{01_section_1.tex}
\input{02_section_2.tex}
\input{03_section_3.tex}
\input{04_section_4.tex}
\input{05_section_5.tex}

{\small
\bibliographystyle{ieee_fullname}
\bibliography{egbib}
}

\newpage
\input{supplementary_material.tex}

\end{document}

%% file: 00_abstract.tex
\begin{abstract}
   The existing solutions for object detection distillation rely on the availability of both a teacher model and ground-truth labels. We propose a new perspective to relax this constraint. In our framework, a student is first trained with pseudo labels generated by the teacher, and then fine-tuned using labeled data, if any available. Extensive experiments demonstrate improvements over existing object detection distillation algorithms. In addition, decoupling the teacher and ground-truth distillation in this framework provides interesting properties such as: 1) using unlabeled data to further improve the student’s performance, 2) combining multiple teacher models of different architectures, even with different object categories, and 3) reducing the need for labeled data (with only 20\% of COCO labels, this method achieves the same performance as the model trained on the entire set of labels). Furthermore, a by-product of this approach is the potential usage for domain adaptation. We verify these properties through extensive experiments.
\end{abstract}

%% file: 01_section_1.tex
\section{Introduction} \label{introduction}

Deployment of deep learning models to edge devices often imposes constraints on size, latency, and runtime memory. Knowledge distillation~\cite{bucilua2006model, hinton2015distilling} is one of the most promising ways of producing compact models.
Knowledge distillation for image classification was introduced in~\cite{hinton2015distilling}, where an ensemble of large teacher models was distilled to a smaller student model without a considerable performance loss. The main idea was to extract the so-called ‘dark knowledge’ and teach that to the student model. This was achieved by introducing a loss term on the ``soft-targets" (more details in Section~\ref{sec:2.1}). Since then, there has been a large amount of publications focused on improving the distillation method of ~\cite{hinton2015distilling} (a.k.a vanilla distillation).

The majority of the distillation related literature (e.g.~\cite{bucilua2006model, hinton2015distilling, romero2014fitnets, wang2018dataset}) focus on the image classification task and in fact the formulation proposed in~\cite{hinton2015distilling} holds only for classification networks. That being said, object detection~\cite{redmon2018yolov3, redmon2016you, girshick2014rich, lin2017focal, ren2015faster, liu2016ssd, li2017light, li2019scale} is a much more practical task, and this motivated others to investigate the distillation for object detection. 

The existing works~\cite{li2017mimicking, wang2019distilling, chen2019analysis, mehta2018object, chen2017learning} on object detection distillation propose to use feature or detector outputs (box, confidence, class probabilities) matching between student and teacher models, so the student’s activations follow those of the teacher’s. These methods are often verified by choosing students of the same architecture as the teacher, but shallower and thinner. These solutions rely on the availability of both a teacher model and ground truth labels. In addition, for multi-teacher distillation, they assume that all teachers detect the same object classes. These assumptions limit the practicality of using knowledge distillation for commercial services where a user uploads a large teacher model to the cloud (but provides no labeled data or a reduced set only), and the goal is to train a compact student model from it. It also makes it difficult to consider the case where teacher models are experts in detecting different type and number of object classes, and potentially with different architectures.

This paper proposes a simple, yet powerful approach for knowledge distillation in object detector neural networks. We argue that knowledge distillation is intrinsically different between the image classification and object detection tasks. In other words, the so called dark knowledge does not lie anymore in some layers’ features. Instead, the student model generalizes better when it is first presented with simpler object labels explicitly. Moreover, student object detectors can incrementally improve from teacher distillation and ground truth training. This idea is in some sense related to~\cite{mirzadeh2019improved} where ‘teacher assistants’ were shown to be helpful for image classification distillation. Teacher assistants introduced in~\cite{mirzadeh2019improved} are neural networks of sizes (thus capacities) between those of the student and the teacher. \cite{mirzadeh2019improved} argued that students learn better first from the teacher assistants, as it is easier for the students to learn logits of a more similar feature space. The fundamental difference between the methodology proposed in this paper (for object detection) and~\cite{mirzadeh2019improved} (for image classification) is that we do not introduce any extra models. Instead, we argue that the student can learn better if it is first trained together with the teacher guidance and a subset of data that is carefully generated for it to learn. In other words, instead of letting the student network learn from the training data on its own, we utilize the teacher to label the data for the student, thereby providing a subset of the whole training dataset, with predictions that are easier to follow by the student.

The proposed framework decouples distilling from the ground truth data and the teacher model from each other. This provides several nice properties such as:
\begin{enumerate}[leftmargin=*, label=\alph*]
  \item Being able to use unlabeled data to further improve the student’s performance. \vspace{-6pt}
  \item Reducing the need for labeled data. \vspace{-6pt}
  \item Distilling from (combining) multiple teacher models of different architectures, even with different object classes (with or without overlap). \vspace{-6pt}
  \item Performing domain adaptation as a by-product, where limited or no labels are available for the target domain.
\end{enumerate}
These properties are often practical necessities for commercial cloud model compression services.  We verify these properties through extensive experiments in Section \ref{section-3}. 

The main contributions of this paper are summarized as:  
\begin{itemize}
    \item \emph{A new object detection distillation strategy}: \\ 
with properties such as being able to improve the distillation performance using unlabeled data, reducing the need for labeled data, or combining multiple detector models. To the best of our knowledge, there has been no previous work on object detection distillation that decouples learning from the teacher and ground truth data, nor one that combines generic object detectors of (non-)overlapping object classes like our method does.
    \item \emph{Insights gained from an extensive set of experiments}:\\
We designed a comprehensive set of experiments to evaluate the proposed object detection distillation algorithm. The experiments are broken down in a way that they provide insights on 1) whether or not using unlabeled data can help the distillation, 2) if so, what is the impact of the unlabeled data size on the performance improvements? 3) the role of techniques such as feature matching, imitation masking, or box matching, in object detection distillation.
\end{itemize}


%% file: 02_section_2.tex
\vspace{-2pt}
\section{Related works} \label{section-2}
\vspace{-2pt}

\subsection{Relations to distillation for image classification} \label{sec:2.1}
Classification distillation exploits soft-targets matching between the teacher and student logits. It was argued in~\cite{hinton2015distilling} that soft-targets provide a better discrimination between the classes with low likelihood values, thereby allowing the student to learn more than just the best target class label:
\begin{equation}
\label{eq:1}
    L = L^{GT} + \alpha L^{KD}
\end{equation}
where $L^{GT}$ denotes the ground truth loss term, $L^{KD}$ is the distillation loss, and $\alpha$ is a weighting parameter.

There are other studies that tried to build upon the baseline distillation of~\cite{hinton2015distilling}. Among them, FitNets~\cite{romero2014fitnets} proposed the student to mimic the teacher feature maps (hints). This imposes a constraint on the student which can sometimes be too strict, since the capacities of teacher and student may differ considerably. Attention Transfer~\cite{zagoruyko2016paying} relaxes the assumptions of FitNet. 
The student network is trained not only to have similar features to the teacher, but also to have similar attention maps. In~\cite{turner2018hakd} authors designed pruned student models and customized the distillation for target hardware. \cite{lopes2017data, chen2019data} proposed data free distillation methods for the classification task where there are no training data available.

Even though many ideas can be borrowed from the classification literature, but there are still some intrinsic differences between the tasks of classification and detection that requires careful specialized designs for object detection distillation. One can still employ soft-logits matching on layers of a detector network (backbone, head, or intermediate layers), but how to distill bounding boxes (a regression problem) may not be best addressed through logits matching.

In addition, classification distillation requires identical target classes between the teacher(s) and student. Different or only partially overlapping classes result in catastrophic learning problems, since unlike detectors (that learn to detect only the target object of interest and ignore everything else), classifiers will have to pick a class anyway. In case a training example doesn't belong to a category known to a teacher, it will be assigned a wrong class. Subsequently, the student will be trained with a wrong class. For example two teachers: one a person/cat classifier, and one a car/bicycle classifier, can't be jointly distilled to a student model (not a problem for detection as we show in Section~\ref{sec:4.4.2}). This problem exists in supervised distillation, and becomes even more challenging in the case of distillation with unlabeled data. Although there have been attempts to address this problem~\cite{vongkulbhisal2019unifying}, but generally its literature is very limited.

That said, supplementary materials \cite{Authors21supplementary} contain more details and numerical evaluations on the classification task.

\subsection{Related works on object detection distillation}

Among the recent works, \cite{li2017mimicking} proposed a framework for distilling knowledge to object detector models where the distillation loss term is based on distance between detector features of the teacher and student. This enforces the student model to mimic the teacher. \cite{wang2019distilling} built on this idea by introducing the concept of imitation (objects) masks. These masks highlight the location of objects. The distillation loss term is masked by the imitation mask, thereby the student is pushed only to learn the objects, without any constraint on the background within the distillation loss term. Note that this algorithm requires the ground truth object labels to be available. On the other hand, \cite{chen2019analysis} argued that in the case of object detector models, the backbone CNN features provide a stronger discriminative ability than the detector head features. Authors in~\cite{chen2019analysis} suggested that students distilled from teacher backbone features generalize better than the ones trained with the detector head features. 

While~\cite{li2017mimicking, wang2019distilling, chen2019analysis} defined knowledge distillation loss as a feature matching distance, \cite{mehta2018object} proposed to directly use the detector outputs (boxes, confidence scores, and class probabilities). To this end, the distillation loss in~\cite{mehta2018object} is defined as three terms on objectness confidence score, class probabilities, and bounding boxes predicted. For each prediction, the two later terms are weighted by the objectness score, so the contribution of each predicted bounding box is according to its confidence level of being an object. It is worth noting that we found in our experiments that this method of knowledge distillation works better when candidate boxes predicted by the teacher model are filtered (e.g. with non-max suppression). Otherwise, if a teacher model is not trained well, its predictions might be too noisy for a student model to learn from. 

A hybrid approach was taken in~\cite{chen2017learning} where distillation loss has a feature matching term in addition to a box/probabilities matching term. The feature matching term is a L2 loss between the backbone features. The detection outputs term however includes a bounded regression loss for the bounding boxes and a cross entropy loss for the classification probabilities.

As mentioned in Section~\ref{introduction}, the existing object detection distillation methods require labeled data to perform distillation, often pose constraints on models architectures, and can't handle non/partially overlapping object categories.

\subsection{Relations to semi/self/un-supervised learning}

There are several lines of works in semi/self/un-supervised learning related to our method. First are the generative models (in the context of distillation are sometimes called data-free). These methods learn to generate realistic examples to train the student with. The examples are generated either from unlabeled training data, or from noise~\cite{yin2020dreaming, brock2018large}. These methods are mostly trained and evaluated on classification datasets and at relatively low resolutions (32x32, 64x64, or highest ones at 256x256~\cite{brock2018large}). On the other hand, state-of-the-art object detectors require high resolutions (e.g. as high as 1536 for EfficientDet-D7~\cite{tan2020efficientdet}). Also, classification methods do not simply extend to detection (see \ref{sec:2.1}). Moreover, generative methods need not only to generate image contents, but also bounding boxes.

Another line of related work is zero-shot distillation, where synthesized data impressions from the teacher are used as surrogates to train the student~\cite{micaelli2019zero, nayak2019zero}. The advantage of these methods is that they do not need any training data. However, similar to generative works, they are mostly in the context of classification (see \ref{sec:2.1}). Moreover, the upper-bound of performance for both these works and generative works is often considered to be the student's performance with full data on supervised training, or on original knowledge distillation~\cite{micaelli2019zero, nayak2019zero, yin2020dreaming, brock2018large}. We go well above these bounds, as shown in Section \ref{section-4}.

In the context of semi-supervised learning, \cite{xie2020self} used a large set of unlabeled data for improving the accuracy of ImageNet classifiers. Note that using unlabeled data in weakly supervised classification (i.e. training with pseudo labels of teacher(s)) is only feasible if the unlabeled data is collected from the same classes of the teacher model. This is due to the fact that each and every training example is assigned to a class, and having images from outside classes results in bad label assignment (less problematic for ImageNet since it has a large number (1000) of classes). For example, one can't use arbitrary images for distilling a cat/dog classifier. This becomes less of an issue when relying on other sources of information, e.g. crawling data from the web, and using meta-data or file tags (but it's not always possible to do so). Our proposed method does not suffer from this phenomenon. It is shown in Section~\ref{sec:4.4.1} that even a two category object detector can effectively be trained via distillation using the entire open images dataset (unlabeled), without a curated example selection procedure.

Self-supervised learning is another related area, that has gained momentum recently. The idea is to learn general representations from one or more auxiliary tasks~\cite{jing2020self}. Since the auxiliary tasks are not aware of the down-stream task, they rely on unsupervised representation learning only. Having a teacher model along side the unlabeled data is expected to perform better, as shown in Section \ref{section-4}. 

%% file: 03_section_3.tex
\section{The proposed method} \label{section-3}
This section elaborates on the proposed knowledge distillation strategy for object detection. Our hypothesis is that in a teacher-student distillation scenario, the student model learns better if it is first trained by a ‘teaching source’ that has a comparable capacity. This makes sense because the intermediate teaching source acts as a hint for the student model during distillation. 


An observation motivated our hypothesis further: if we allow the student model to learn only the teacher’s understanding of the data, it can later generalize better on the whole set of ground truth labels. To this end, we label the training data with a pre-trained teacher model. Since the teacher model does not have a perfect detection ability, it will only detect a limited number of object instances within a subset of training samples. These samples are the ones that are likely easier for the student to learn from and teacher detections (although they could be noisy) are the ones that the student can follow better than trying to explore the entire search space on its own. This was tested with a two class subset of the Microsoft COCO dataset~\cite{lin2014microsoft} (Person \& Bicycle). After learning from the teacher's pseudo labels, fine-tuning was done using the ground truth labels. The 2-stage training strategy achieved a higher mean Average Precision (mAP) value than the standard supervised training of the student (See Section \ref{section-3} for more details and a comprehensive set of follow-up experiments).


\begin{algorithm}
    \HandRight { Dataset $D$: Gather a large set of images, they do not
    
    \hspace{12pt} have to have labels.}
    
    \HandRight { Have a pre-trained teacher model perform a forward pass
    
    \hspace{10pt} on dataset $D$. Collect the predictions.}
    
    \HandRight{ Perform distillation according to (\ref{eq:2}) on $D$: 
    
    \hspace{10pt} One loss term for matching the features with imitation 
    
    \hspace{10pt} masking, and another term for matching the teacher and
    
    \hspace{10pt} student predictions.}
    
    \HandRight{ End here if no additional labeled data are available.}
    
    \HandRight{ Otherwise, fine-tune the student over the labeled data.}

    \caption{Knowledge distillation for object detection.}
\end{algorithm}

\begin{figure*}
    \centering
    \includegraphics[width=1.0\linewidth]{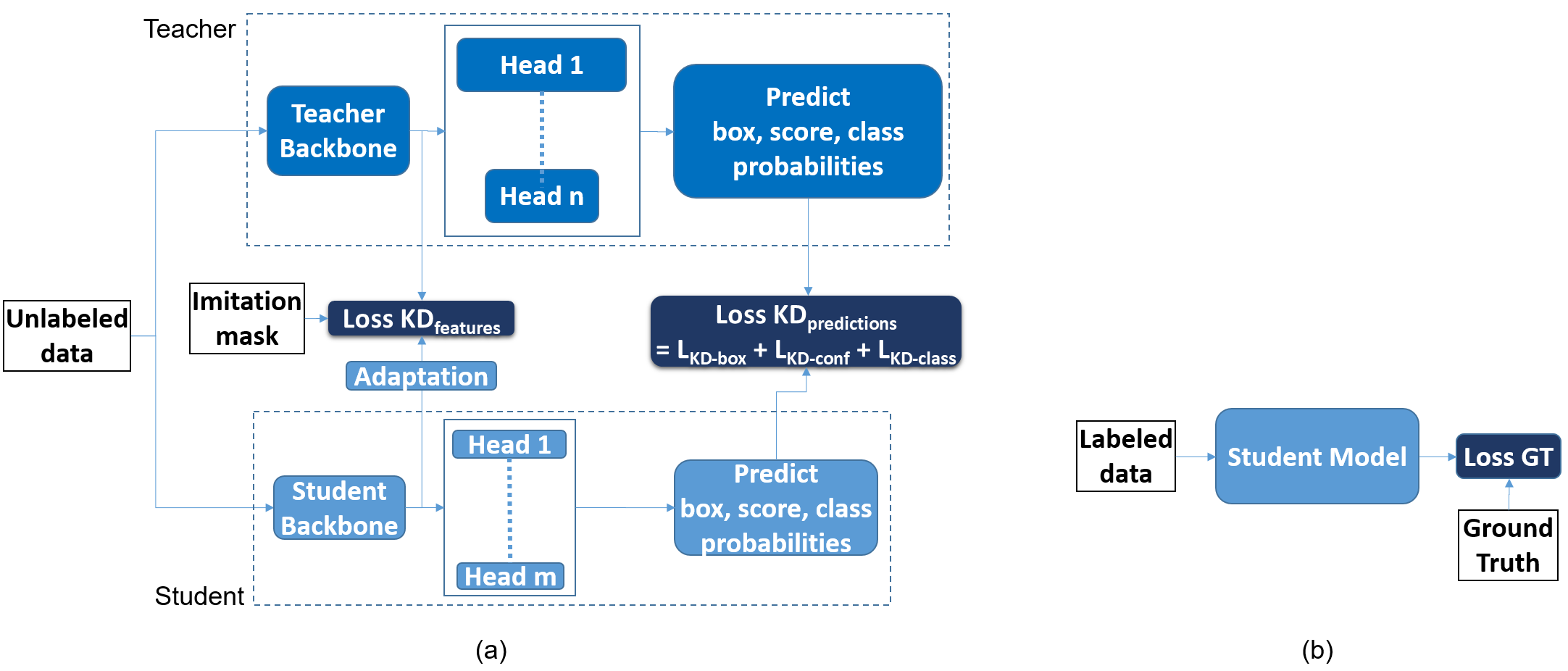}\vspace{-8pt}
    \caption{Distillation for object detection: (a) Step 1: distilling from only a teacher model, and (b) Step 2: fine-tuning using any available labeled data.}
    \label{fig:proposed_method}
\end{figure*}

The observation above further motivates the proposed distillation framework. In this framework, the student model is trained in two steps: 1) distillation using the teacher model only, without seeing the ground truth training labels, and 2) fine-tuning over the ground truth training data. Decoupling the distillation in this way allows the student model to first learn from a teacher-guided sub-space of the original parameter space. This is an easier job for the student and positions it to learn better when it sees the ground truth labels during the fine-tuning step. Fig. \ref{fig:proposed_method} illustrates the architecture of this approach.

Loss function for the first distillation step is defined as: 
\begin{equation}
\label{eq:2}
    L^{KD} = L_F^{KD} + L_P^{KD}    
\end{equation}
where $L^{KD}$ refers to the distillation loss. $L_F^{KD}$ and $L_P^{KD}$ denote the feature matching and object detection prediction components of the distillation loss. $L_F^{KD}$ is define as:
\begin{equation}
\label{eq:3}
    L_F^{KD} = || F^T \otimes M^T - F^S \otimes M^T  || _2^2
\end{equation}
where $F^T$ and $F^S$ are features from the teacher and student, $M^T$ denotes the imitation (objects) mask generated from the bounding box predictions of the teacher, and $\otimes$ is the element-wise multiplication operator. Our experiments were in agreement with~\cite{chen2019analysis} in that backbone features prove to be more useful. Also, the imitation (objects) masks~\cite{wang2019distilling} are being applied to the backbone features (not the detector heads). In order to ensure the shapes and dimensions are compatible for matching, an adaptation block needs to be added after the students feature response. We found out in our experiments that a minimal adaptation size achieves the best performance, so a one layer convolution was used for adaptation. This makes sense since only the main student network without the adaptation is later used for validation, so the knowledge of detection should stay in the actual student model and not the adaptation piece. It is also worth noting that Non-Max Suppression (NMS) is applied to the teacher model predictions to reduce the noisiness of the predicted bounding boxes.
In the case of YOLO-based object detectors~\cite{redmon2016you}, the prediction loss component is defined as:
\begin{equation}
\label{eq:4}
    L_P^{KD} = L_{box}^{KD} + L_{conf}^{KD} + L_{class}^{KD}    
\end{equation}
The three components in (\ref{eq:4}) account for bounding box regression, objectness confidence, and class probability. We used a modified loss definition compared to the original YOLO model~\cite{redmon2016you} for better convergence.

The loss components in our set up are defined as:
\begin{equation}
\label{eq:5}
    L_{box}^{KD} = L_{xy}^{KD} + L_{wh}^{KD}    
\end{equation}
\begin{equation}
\label{eq:6}
    L_{xy}^{KD} = \sum_{i=0}^{K^2}\sum_{j=0}^{B}
    M_{ij}^T [(x_{ij}^T - x_{ij}^S )^2 + (y_{ij}^T-y_{ij}^S )^2]    
\end{equation}
\begin{equation}
\label{eq:7}
    L_{wh}^{KD} = \sum_{i=0}^{K^2}\sum_{j=0}^{B}
    M_{ij}^T [(w_{ij}^T - w_{ij}^S)^2 + (h_{ij}^T - h_{ij}^S)^2]    
\end{equation}
\begin{equation}
\label{eq:8}
    \begin{split}
        L_{conf}^{KD} = \sum_{i=0}^{K^2} & \sum_{j=0}^{B}
        M_{ij}^T \times \sigma_E (M_{ij}^T,C_{ij}^S) \\ 
        & + (1-M_{ij}^T) \times \mathbbmss{1}_{ij}^{ign} \times \sigma_E (M_{ij}^T,C_{ij}^S)
    \end{split}
\end{equation}
\begin{equation}
\label{eq:9}
    L_{class}^{KD} = \sum_{i=0}^{K^2} \sum_{j=0}^{B}
    M_{ij}^T \times \sigma_E (M_{ij}^T,P_{ij}^S)
\end{equation}
where $L_{xy}^{KD}$ and $L_{wh}^{KD}$ are regression losses for the center and size of the boxes, $B$ is number of predicted boxes, $K$ is number of YOLO grid cells in each direction, $M_{ij}^T$ is object (imitation) mask defined by the teacher at coordinate location $(i,j)$, $(x,y)$ is center of a box, $(w,h)$ are width and height of a box, $\mathbbmss{1}_{ij}^{ign}$ is an ignore mask (0 when IOU between the predicted box and the teacher’s box is less than a threshold e.g. 0.5, and 1 otherwise), $C_{ij}^S$ is confidence logit predicted by the student, $\sigma_E$ is sigmoid cross entropy, and $P_{ij}^S$ denotes class probability logits predicted by the student.

In the case of YOLO architectures with more than one scale (e.g. YOLOv3), the distillation loss defined in (\ref{eq:2}) is calculated per each scale and then summed up to form the overall loss. Also, in the case of RCNN like detectors~\cite{girshick2014rich} the formulation of (\ref{eq:4}) needs to be modified accordingly.

The second step of distillation leverages the ground truth labels. Loss is defined similar to (\ref{eq:4}), but between the student predictions and the ground truth:
\begin{equation}
\label{eq:10}
    L^{GT} = L_{box} + L_{conf} + L_{class}    
\end{equation}
where $L^{GT}$ is the detection loss between the student predictions and the ground truth labels. The three loss components of (\ref{eq:10}) are calculated similar to (\ref{eq:5})-(\ref{eq:9}). However, instead of the teacher model predictions, the ground truth labels are used. During the optimization, $L^{KD}$ and $L^{GT}$ are minimized independently, one after the other.

This way of knowledge distillation provides several interesting characteristics: \vspace{-2pt}

\begin{enumerate}[leftmargin=*, label=\alph*]
  \item \textbf{Usage of unlabeled data}: Since the first distillation step does not use any ground truth labels, it is possible to leverage an arbitrary large set of unlabeled images. Teacher’s knowledge is distilled to student over a large dataset and this can results in a more accurate student. \vspace{-4pt}
  \item \textbf{Reducing the need for labeled data}:  Data labeling is costly. A model that is trained through distillation by a teacher (and potentially large amounts of unlabeled data), may only need a limited amount of additional labels to achieve an acceptable performance (Section \ref{section-3}). \vspace{-12pt}
  \item \textbf{Combining (merging) pre-trained teacher models}: To this end, different teachers go over the data and provide their predictions. The predictions are then aggregated. The student model is trained with the resulting data/labels to achieve a fair performance on the union of all teachers object classes. Having labeled data for fine-tuning can boost the student model’s performance. \vspace{-2pt}
  \item \textbf{Domain adaptation via distillation}, i.e. to distill knowledge of the teacher’s domain to a student in another domain: Suppose teacher $T$ is trained with data in domain $D^T$. The goal is to train a student model to work with data in domain $D^S$ that is similar to $D^T$ but not identical. An example application would be a surveillance camera system where day-time data from one camera are labeled and a model is well trained on those data (teacher). A second camera (student) that operates at night has no labeled data. It is possible to leverage the teacher-student framework proposed here to distill the knowledge from the day-time camera to the night-time one. To this end, the teacher model is used to make predictions on data collected from the student domain, so the student can be trained on these predictions. If any labeled data are available from the student domain they can be used for fine-tuning the student, otherwise, even fine-tuning on the teacher domain helps improving the student’s model performance on the student domain data (More details in Section \ref{section-3}).
\end{enumerate}

\vspace{-4pt}
As mentioned earlier, the four properties above are important for supporting practical use-cases and applications of knowledge distillation. In the next section, we provide experiments results for each of these cases.

%% file: 04_section_4.tex
\section{Experiment results and discussions} \label{section-4}
\vspace{-2pt}
\subsection{Model architectures}
Without the loss of generality, we choose YOLOv3 architecture for most of experiments. Note that the distillation framework proposed in this paper fits well also with other detector models such as RCNN or RetinaNet based detectors. In most of the experiments, the original YOLOv3 architecture with DarkNet53 backbone is selected as the teacher. Later in Section \ref{section-3} we also try a FasterRCNN teacher to study the effect of architecture change. For the student, we use a custom slim YOLOv3 with a 23 layer backbone. This model is shallower and thinner than the teacher, and its size on disk is around one tenth of the teacher. Training is done from scratch with ImageNet pre-trained weights. Hyper-parameters such as non-max suppression parameters, score threshold, maximum number of detections, and other parameters are consistent between the teacher and student, and are similar to the ones used in the DarkNet implementation of YOLOv3~\cite{redmon2018yolov3}.

\subsection{Tuning and training tricks}
When evaluating the distillation performance, we need to ensure that the student, teacher, and distilled student models are highly tuned and have achieved a capacity that may not be improved further with common tuning and training tricks. To this end, we employed the following techniques to boost the validation accuracy for the trained models. With that, we tried to decouple improvements that can be made through distillation from teacher models and the ones from typical training tricks and tunings:
\vspace{4pt}

\noindent \textbf{Augmentation}:

•	Random crop (with constraints on bounding box)

•	Random color distortions in the HSV domain

•	Random flip: horizontal or vertical

•	Random expansion (place to a larger canvas)
\vspace{4pt}

\noindent \textbf {Multi-scale input}:
Performance of object detectors is usually compared at a given input image resolution. Using various input sizes (in the range of $320\times320$ to $640\times640$) dynamically during the training consistently improved the performance in our experiments. The validation image size is set at $416\times416$.

\vspace{4pt}

\noindent \textbf {Mix-up training}:
Mix-up was first introduced for classifiers and GANs~\cite{zhang2017mixup} to alleviate issues such as memorization and sensitivity to adversarial examples. It involves with mixing training examples and their associated labels. This idea was modified here to be used for the detection task.
\vspace{4pt}

\noindent \textbf {Label smoothing}:
Label smoothing did not help with the distillation performance and is not used in our experiments. This finding is in agreement with the state-of-the-art~\cite{muller2019does}.
\vspace{4pt}

\noindent \textbf {Focal loss}:
Focal loss~\cite{lin2017focal} was initially proposed to improve the object detection performance on hard examples and to address the class imbalance between foreground and background. We include focal loss during the training.
\vspace{4pt}

\noindent \textbf {Learning rate scheduling}:
For each training job, we tried separately a wide range of learning rate scheduling methods: exponential, piecewise, fixed, cosine decay, and cosine decay with restart. We also used a warm up~\cite{gotmare2018closer} strategy as it prevents divergence at the start of training.

\subsection{Evaluation metric}
After each training job, the model with the best validation mAP is selected. The training process is repeated for 10 times to reduce the impact of random initialization on the performance measured (around 2-3K GPU-hours per each table entry reported here). The average mAP along with the range of mAP for best models are presented in this section. We utilized the mAP implementation available in the FAIR’s Detectron repo~\cite{Detectron2018}. The mAPs @50\% performance are reported in this section.

\subsection{Performance evaluations}
\subsubsection{Teacher-Student distillation experiments}\label{sec:4.4.1}
The proposed teacher-student framework is evaluated on both small and large scale number of object classes to ensure the results do not behave differently at various scales. We first evaluate the models using a subset of COCO dataset with only two object classes, ‘person’ and ‘bicycle’. Then we evaluate the models on the entire COCO 2017 object detection dataset (80 classes, 118K training and 5K validation examples). Moreover, later in this section we design experiments in which multiple two-class teacher object detectors are combined to form a multi-class student.

Table \ref{tab:2} shows the performance on two class object detection. For distilled models in this table, we used the COCO dataset, the subset images that include at least one of these two classes. Table \ref{tab:3} shows the results when OpenImagesV5 dataset~\cite{krasin2017openimages} is used for teacher only distillation step (out of 1.7M images, we use whatever many the teacher can predict at least one bounding box on). No labels were available for this step. In all experiments, the validation set of COCO was used for mAP calculation. Also, the tables in this section incorporate abbreviations when referring to various methods, for the sake of brevity. To this end, SD, FM, IM, PM, UD, FT, SSL, and OID denote "Supervised Distillation", "Feature Matching", "Imitation Masking", "Predictions (boxes, scores, and classes) Matching", "Unsupervised Distillation (only distill with teacher's pseudo labels on unlabeled data)", "Fine-Tuning", "Self-Supervised Learning", and Open Images Dataset, respectively.

It is observed from Table \ref{tab:2} that the proposed distillation strategy performs well. Feature matching and imitation masks have also improved the performance. In addition, Table \ref{tab:3} shows that using unlabeled data has considerably improved the overall mAP. Supplementary material~\cite{Authors21supplementary} contains examples were students trained with distillation can detect a higher number of bounding boxes than the students trained supervised without distillation.

To study the case that teacher and student are from different architectures and trained on different data, we change the teacher to a FasterRCNN model trained on Open Images dataset. Table \ref{tab:4} shows results for this experiment, where the FasterRCNN teacher is distilled to the custom YOLO-based student. It is observed from this table that changing the architecture slightly reduces the improvement gap, but the trend is consistent with Table \ref{tab:2} and \ref{tab:3}.

\begin{table}
    \centering
    \caption{Two class knowledge distillation. (63K training samples, 2.7K validation). Abbreviations defined in  \ref{sec:4.4.1}.}\vspace{-6pt}
    \setlength{\tabcolsep}{0.01em}
    \renewcommand{\arraystretch}{1.0}%
    \begin{tabular}{|C{5.4cm}|C{2.85cm}|}
    \hline
        Model & Mean mAP \small{over 10 runs (min, max)} \\
    \hline \hline
        Teacher & 0.6009 \small{(0.599,0.603)} \\ \hline
        Baseline: student supervised & 0.5295 \small{(0.527,0.532)} \\ \hline \hline
        SD\textsubscript{COCO-FM-IM}~\cite{wang2019distilling} & 0.5421 \small{(0.541,0.544)} \\ \hline
        SD\textsubscript{COCO-FM}~\cite{li2017mimicking} & 0.5365 \small{(0.534,0.539)} \\ \hline
        SD\textsubscript{COCO-PM}~\cite{mehta2018object} & 0.5284 \small{(0.527,0.531)} \\ \hline \hline 
        UD\textsubscript{COCO} & 0.4553 \small{(0.453,0.458)} \\ \hline
        UD\textsubscript{COCO} + FT\textsubscript{COCO} & \textbf{0.5533} \small{(0.553,0.554)} \\ \hline \hline
        UD\textsubscript{COCO-FM-IM} & 0.4673 \small{(0.466,0.468)} \\ \hline
        \textbf{Ours}: UD\textsubscript{COCO-FM-IM} + FT\textsubscript{COCO} & \textbf{0.5629} \small{(0.562,0.564)} \\
    \hline
    \end{tabular}
    \label{tab:2}
\end{table}

\begin{table}
    \centering
    \caption{Two-class object detection distillation using larger unlabeled set (63K labeled + 1.03M unlabeled training samples, 2.7K validation). Abbreviations are defined in~\ref{sec:4.4.1}.}\vspace{-6pt}
    \setlength{\tabcolsep}{0.01em}
    \renewcommand{\arraystretch}{1.0}%
    \begin{tabular}{|C{5.4cm}|C{2.85cm}|}
    \hline
        Model & Mean mAP \small{over 10 runs (min, max)} \\ \hline \hline
        UD\textsubscript{OID} & 0.4940 \small{(0.492,0.496)} \\ \hline
        UD\textsubscript{OID} + FT\textsubscript{COCO} & 0.5815 \small{(0.580,0.583)} \\ \hline \hline
        UD\textsubscript{OID-FM-IM} & 0.4993 \small{(0.498,0.501)} \\ \hline
        \textbf{Ours}: UD\textsubscript{OID-FM-IM} + FT\textsubscript{COCO} & \textbf{0.5899} \small{(0.589,0.591)} \\ 

    \hline
    \end{tabular}
    \label{tab:3}
\end{table}

\begin{table}
    \centering
    \caption{Using an entirely different teacher architecture (FasterRCNN teacher to a slim YOLO-based student) (63K labeled + 1.03M unlabeled training samples, 2.7K validation). Abbreviations are defined in~\ref{sec:4.4.1}.}\vspace{-6pt}
    \setlength{\tabcolsep}{0.01em}
    \renewcommand{\arraystretch}{1.0}%
    \begin{tabular}{|C{5.4cm}|C{2.85cm}|}
    \hline
        Model & Mean mAP \small{over 10 runs (min, max)} \\
    \hline \hline
        UD\textsubscript{COCO} & 0.4248 \small{(0.422,0.426)} \\ \hline
        UD\textsubscript{COCO} + FT\textsubscript{COCO} & 0.5425 \small{(0.540,0.545)} \\ \hline \hline
        UD\textsubscript{OID} & 0.4592 \small{(0.456,0.463)} \\ \hline
        \textbf{Ours}: UD\textsubscript{OID} + FT\textsubscript{COCO} & \textbf{0.5647} \small{(0.562,0.566)} \\ 
    \hline
    \end{tabular}
    \label{tab:4}
\end{table}

\begin{table}
    \centering
    \caption{Object detection distillation on COCO dataset (80 classes) (118K labeled + 1.28M unlabeled training samples, 5K validation). Abbreviations are defined in~\ref{sec:4.4.1}.} \vspace{-6pt}
    \setlength{\tabcolsep}{0.01em}
    \renewcommand{\arraystretch}{1.0}%
    \begin{tabular}{|C{5.4cm}|C{2.85cm}|}
    \hline
        Model & Mean mAP \small{over 10 runs (min, max)} \\
    \hline \hline
        Teacher & 0.6206 \small{(0.618,0.623)} \\ \hline
        Baseline: Student supervised & 0.3449 \small{(0.342,0.347)} \\ \hline 
        SSL\textsubscript{COCO-rotnet} + FT\textsubscript{COCO}~\cite{gidaris2018unsupervised} & 0.3303 \small{(0.330,0.331)} \\ \hline
        SSL\textsubscript{OID-rotnet} + FT\textsubscript{COCO}~\cite{gidaris2018unsupervised} & 0.3429 \small{(0.342,0.343)} \\ \hline
        SD\textsubscript{COCO-FM-IM}~\cite{wang2019distilling} & 0.3817 \small{(0.381,0.383)} \\ \hline
        SD\textsubscript{COCO-FM}~\cite{li2017mimicking} & 0.3755 \small{(0.375,0.376)} \\ \hline
        SD\textsubscript{COCO-PM}~\cite{mehta2018object} & 0.3694 \small{(0.369,0.370)} \\ \hline \hline
        UD\textsubscript{COCO} & 0.3435 \small{(0.341,0.346)} \\ \hline
        UD\textsubscript{COCO} + FT\textsubscript{COCO} & 0.4015 \small{(0.400,0.403)} \\ \hline \hline
        UD\textsubscript{COCO-FM-IM} & 0.3496 \small{(0.348,0.351)} \\ \hline
        UD\textsubscript{COCO-FM-IM} + FT\textsubscript{COCO} & 0.4079 \small{(0.406,0.409)} \\ \hline \hline
        UD\textsubscript{OID} & 0.3567 \small{(0.355,0.359)} \\ \hline
        UD\textsubscript{OID} + FT\textsubscript{COCO} & 0.4166 \small{(0.415,0.418)} \\ \hline \hline
        UD\textsubscript{OID-FM-IM} & 0.3608 \small{(0.359,0.361)} \\ \hline
        \textbf{Ours}: UD\textsubscript{OID-FM-IM} + FT\textsubscript{COCO} & \textbf{0.4220} \small{(0.421.0.423)} \\ 
    \hline
    \end{tabular}
    \label{tab:5}
\end{table}

\begin{figure}
    \centering
    \includegraphics[width=1.0\linewidth]{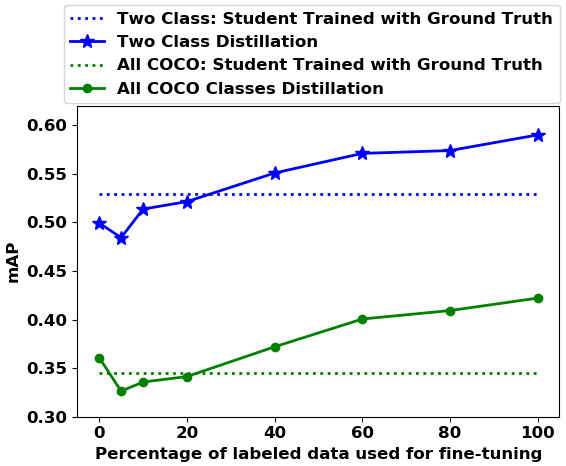} \vspace{-8pt}
    \caption{Distillation with limited amounts of labeled data}
    \label{fig:reduced_labeled_set}
\end{figure}

In the case of complete (80 object classes) COCO dataset, similar conclusions can be drawn, as observed in Table \ref{tab:5}. Moreover, Fig. \ref{fig:reduced_labeled_set} shows the mAP achieved when the size of labeled data for the second step of distillation (fine-tuning) is reduced. It is observed from Fig. \ref{fig:reduced_labeled_set} that even with 20\% of labels, distilled student is on par with the student trained with all labels. Note that if the data size used for fine-tuning is too small and the student is fine-tuned for long, it will then over-fit to the small set of fine-tuning data, and yields lower validation accuracy.

An interesting observation in Fig. \ref{fig:reduced_labeled_set} is that in the 80-class case, the student trained only with the teacher without any fine-tuning can outperform the student that is trained supervised with all labels. This is not the case for the two-class scenario. The reason is that the 80-class teacher is much more accurate than the 80-class student trained from scratch. There is around 28\% mAP gap between the two, while this gap for the two-class case is around 7\%. That means the teacher’s knowledge in the 80-class case becomes much more valuable for the student, and with the help of unlabeled data it can achieve a reasonable performance. In the two-class case however, the student does not rely on the teacher as much, and generally has an easier task to learn.

\subsubsection{Combining object detectors} \label{sec:4.4.2} \vspace{-2pt}
The proposed knowledge distillation framework allows for ability to merge multiple object detectors, even when they have overlapping object classes. To this end, all teachers perform a forward pass on a set of (unlabeled) examples and their predictions are collected and aggregated. The student is then trained on this collection with the distillation formulation of (\ref{eq:2}). Note that the proposed approach doesn't assume any constraints on the type and number of object categories associated to different object detectors that are to be combined. Therefore, feature maps corresponding to different detectors (with potentially different object categories) highlight spatially different regions. As a result, it doesn't make sense anymore to apply feature matching or imitation masking to these detectors. Consequently, the feature matching loss term on equation (\ref{eq:2}) is discarded for this experiment. After unsupervised distillation, the student is fine-tuned with labeled data, if any are available. It is also worth noting that in general, aggregating the predictions of different models can be done in various ways such as a) affirmative: stacking all predictions and considering them all (even with duplicates), b) consensus: More than half of the teacher models must agree to consider that a region contains an object (based on IOU), and c) unanimous: All teacher models must agree to consider that a region contains an object~\cite{casadoensemble}. In our case, where teachers can have non-overlapping object categories, only the affirmative strategy makes sense. In addition, after the aggregation we can optionally perform NMS to reduce the overlapping predictions. We noticed that applying NMS does not result in an improvement in the combined model's performance, likely due to the fact that noisy predictions are helping the student model to learn better. To verify this solution, we designed an experiment in which there are 3 teacher models, each detecting 2 object classes from the COCO dataset. Table \ref{tab:6} shows the results of this experiment, and confirms the effectiveness of the proposed solution. Supplementary materials \cite{Authors21supplementary} contain additional results.

\begin{table}
    \centering
    \caption{Learning from multiple teachers (70K (labeled) + 1.31M (unlabeled) training samples, 3K validation). Abbreviations are defined in section~\ref{sec:4.4.1}.}\vspace{-8pt}
    \setlength{\tabcolsep}{0.01em}
    \renewcommand{\arraystretch}{1.0}%
    \begin{tabular}{|C{5.4cm}|C{2.85cm}|}
    \hline
        Model & Mean mAP \small{over 10 runs (min, max)} \\
    \hline \hline
        Teacher 1: Person-Car & 0.6668 \small{(0.665,0.669)} \\ \hline
        Teacher 2: Car-Cat & 0.7299 \small{(0.728,0.732)} \\ \hline
        Teacher 3: Person-Bicycle & 0.6009 \small{(0.599, 0.603)} \\ \hline
        Student: Person-Bicycle-Car-Cat & 0.5844 \small{(0.582,0.586)} \\ \hline \hline
        UD\textsubscript{COCO} (3 teachers) & 0.5250 \small{(0.524,0.527)} \\ \hline
        UD\textsubscript{COCO} (3 teachers) + FT\textsubscript{COCO} & 0.6008 \small{(0.598,0.602)} \\ \hline \hline
        UD\textsubscript{OID} (3 teachers) & 0.5386 \small{(0.536,0.540)} \\ \hline
        \textbf{Ours}: UD\textsubscript{OID} (3 teachers) + FT\textsubscript{COCO} & \textbf{0.6269} \small{(0.624,0.629)} \\
    \hline
    \end{tabular}
    \label{tab:6}
\end{table}

\vspace{-4pt}
\subsubsection{Domain adaptation via knowledge distillation} \label{sec:4.4.3} \vspace{-2pt}
The proposed distillation framework holds in case teacher and student models are intended for two slightly different domains. In this case, knowledge from the teacher’s domain is transferred to the student domain. Suppose teacher is trained on dataset $A$, in our experiment a subset of COCO dataset that contains one or more humans captured during the day. The student is supposed to learn to detect humans in a different subset of COCO, dataset $B$ that is non-overlapping with dataset $A$, and contains images of people captured at night. 


Distillation process starts with teacher model to train on $A$, and then perform a forward pass on $B$ to collect its labels. Teacher predictions are distilled to student according to the proposed distillation approach. If no labels are available from $B$, then the student is fine-tuned over $A$, otherwise it is fine-tuned on B. It is observed in Table \ref{tab:7} that distillation improves the adaptation performance.

We provide additional results on the task of robustness against corruptions (another form of domain shift) in the supplementary materials \cite{Authors21supplementary}. Future works include applying the proposed method to classical domain adaptation datasets such as SVHN $\leftrightarrow$ MNIST or KITTI $\leftrightarrow$ Cityscapes.

\begin{table}
    \centering
    \caption{Knowledge distillation for domain adaptation (2K day \& 2K night images for training, 1K night images for validation). Abbreviations are defined in section~\ref{sec:4.4.1}.}\vspace{-8pt}
    \setlength{\tabcolsep}{0.01em}
    \renewcommand{\arraystretch}{1.0}%
    \begin{tabular}{|C{5.4cm}|C{2.85cm}|}
    \hline
        Model & Mean mAP \small{over 10 runs (min, max)} \\
    \hline \hline
        Baseline: teacher trained on day data & 0.3325 \small{(0.330,0.334)} \\ \hline \hline
        UD\textsubscript{night} & 0.2618 \small{(0.258,0.263)} \\ \hline
        UD\textsubscript{night} + FT\textsubscript{day} & 0.3771 \small{(0.375,0.379)} \\ \hline
        UD\textsubscript{night} + FT\textsubscript{night} & 0.4318 \small{(0.429,0.434)} \\ \hline \hline
        UD\textsubscript{night-FM-IM} & 0.2779 \small{(0.276,0.281)} \\ \hline
        UD\textsubscript{night-FM-IM} + FT\textsubscript{day} & 0.3972 \small{(0.394,0.399)} \\ \hline
        \textbf{Ours}: UD\textsubscript{night-FM-IM} + FT\textsubscript{night} & \textbf{0.4578} \small{(0.455,0.459)} \\
    \hline
    \end{tabular}
    \label{tab:7}
\end{table}

%% file: 05_section_5.tex
\section{Conclusion} \label{section-5}
This paper proposes a new perspective on knowledge distillation for object detection. It proposes to decouple the distillation from teacher and labeled data. To this end, the teacher model uses a pool of unlabeled data to provide the student with a subset of the entire parameters space to search in. The teacher distillation uses feature matching with imitation masking and detection loss on bounding boxes, confidence scores, and class probabilities. If any labeled data are available it will be used for fine-tuning the student. This way of distillation allows for leveraging unlabeled data, combining several teacher models with different object classes, and seems promising for domain adaptation.
\balance

%% file: supplementary_material.tex
\section{Supplementary materials}
This section contains additional experiments results and ablation studies.

\subsection{Classification results}
The focus of the paper is on object detection distillation. As mentioned in section ~\ref{section-2}, classification distillation, although has some similarities to detection distillation, but there are major differences that make the impact of distillation with unlabeled data quite different on the two cases: 1) bounding boxes are learned as a regression problem, and may not be best found by feature matching (like it is done for classification distillation). 2) for classification distillation, unlabeled examples need to belong to the categories of the teacher classifier, otherwise they will be assigned with misleading pseudo-labels. This is less of a problem when testing with a large number of classes such as ImageNet-1000 and unlabeled examples from the web, since likely most of the unlabeled examples will belong to one of those 1000 classes. However, this will be a significant problem for classifiers with a low number of classes. 3) When distilling with more than one classifier teachers, they have to have identical target classes, otherwise in case of disjoint or partially overlapping teacher classifiers, the student with be distilled with meaningless aggregated predictions.

That being said, we include experiments for classification distillation with unlabeled examples, to study the differences mentioned above, and get a quantitative sense. To this end, we select ResNet-50 architecture for the teacher, and ResNet-18 for the student. As baselines, we train the student and teacher over Caltech-256 classification dataset, with pretrained ImageNet initializations. The unlabeled data is a 150K subset selected from the OpenImages dataset (OID), but only the examples that belong to the target classes of the Caltech-256 dataset (see the previous paragraph for the reason). We compare the proposed method simplified to classification, against supervised training of the student, as well as vanilla knowledge distillation, and a hybrid distillation (vanilla distillation on OID unlabeled subset, then finetune with labeled Caltech dataset). 

\begin{figure} [ht]
    \centering
    \includegraphics[width=1.0\linewidth]{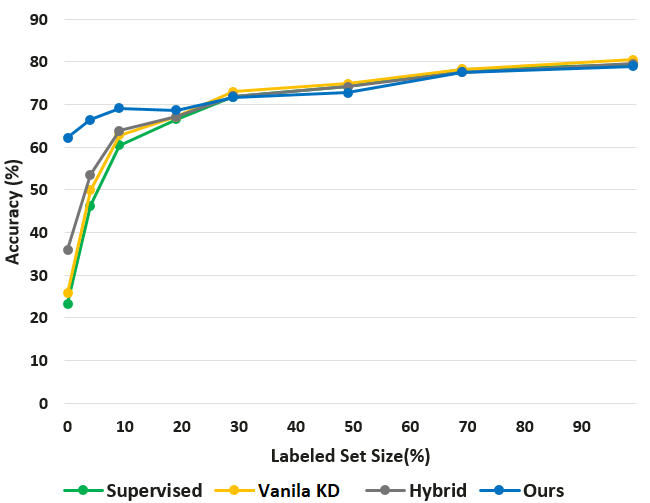}
    \caption{Top-1 results for classification distillation.}
    \label{fig:classification_distillation}
\end{figure}

Figure ~\ref{fig:classification_distillation} shows the results for the classification distillation experiments, when none or partial labeled data are available. It is observed from this figure that the proposed method significantly benefits from unlabeled data compared to other baselines when no labels are available. However, as the portion of the labeled data for finetuning increases, the gains tend to diminish. This observation is consistent with our intuitions and previous explanations that detection generally benefits more from this method than classification. Moreover, our results showed that the student trained with our method can become $5\times$ faster than the teacher when it is also quantized to 8-bits weights/activations ($3\times$ speed-up without quantization).

\subsection{More detection results - Pascal-VOC}
We performed object detection distillation experiments also on the Pascal-VOC dataset, to study the generality of our method on an additional dataset. The baseline is supervised training on VOC07+12 labeled training set. For unlabeled data, we selected a 60K subset of the OpenImages dataset (OID) since Pascal-VOC contains relatively a low number of examples and even at 60K unlabeled example we still see good results. Increasing the size of the unlabeled set will likely improve our results. For our method, we performed `UD\textsubscript{OID-FM-IM} + FT\textsubscript{VOC}' i.e. our method in full, both unsupervised distillation with feature matching and imitation masking, and supervised finetuning. For finetuning we examined 0, 10, and 100 \% availability of labeled examples. Table ~\ref{tab:sup_1} shows the results of this experiment. Results in this table are consistent with our previous observations, and show an advantage when using our method.

The student model is $6\times$ smaller in size than the teacher, $2\times$ faster in floating point, and $3\times$ when quantized.

\begin{table} [ht]
    \centering
    \caption{Object detection distillation on Pascal-VOC dataset (VOC07+12 labeled training set + 60K unlabeled training samples from OID, VOC07 test set).}\vspace{-6pt}
    \setlength{\tabcolsep}{0.01em}
    \renewcommand{\arraystretch}{1.0}%
    \begin{tabular}{|C{2.2cm}|C{2cm}|C{2cm}|C{2cm}|}
    \hline
        Model/mAP & 0\% labels & 10\% labels & 100\% labels\\ \hline \hline
        Supervised & NA & 14 \% & 65.43 \%\\ \hline
        Ours & 48.73\% & 54.64 \% & \textbf{67.2} \% \\
    \hline
    \end{tabular}
    \label{tab:sup_1}
\end{table}

\newpage
\subsection{Improving robustness against corruptions}
\cite{hendrycks2020many, michaelis2019benchmarking, sohn2020simple} reported that when a model is trained on clean data, but tested on corrupt data, its performance will go down. This is because the test set has a somewhat shifted distribution. We studied a small-scale domain shift experiment in section \ref{sec:4.4.3}. Here we investigate the effectiveness of the proposed method for addressing the robustness for corruptions. To this end, we evaluate the performance of the proposed method on Pascal-VOC-C, which contains 15 types of realistic corruptions added to the test set only (See \cite{hendrycks2020many, michaelis2019benchmarking, sohn2020simple} for more details on corruptions). Same as before, teacher a YOLOv3 and student is a very narrow and shallow custom version of it. It is observed from Table \ref{tab:sup_2} that the proposed method can improve the robustness against the synthetic corruptions.

\subsection{More results on combining detector models}
We include an experiment where two object detectors with different architectures are trained on different datasets, and are to be combined via the proposed method. To this end, we consider an EfficientDet-D0 model trained on Pascal-VOC, and an EfficientDet-D1 trained on COCO. The objective is to distill these teachers to yet another different architecture, e.g. a RetinaNet with ResNet-50 backbone. This way, we can study the impact of having teachers with different architectures and training sets, on a student with different architecture than the teachers, distilled on the union of the teachers' object categories.

Table \ref{tab:sup_3} shows the results of this experiment. The proposed method shows improvements compared to supervised training with ground truth labels. An interesting observation here is that for the unsupervised distillation step, when using the combination of COCO and VOC (training sets of teachers), we achieve higher gains than using OID. The intuition is that the small EffDet-D0 model trained on only the VOC dataset, is not good enough to generate quality pseudo lables on the OID dataset. Nevertheless, in both cases our method shows better results than supervised training with 100\% of the labels. Future work includes experimenting with other datasets than VOC, with a different object category set than COCO.

\subsection{Additional demonstrations}
\begin{figure*}
    \centering
    \includegraphics[width=1.0\linewidth]{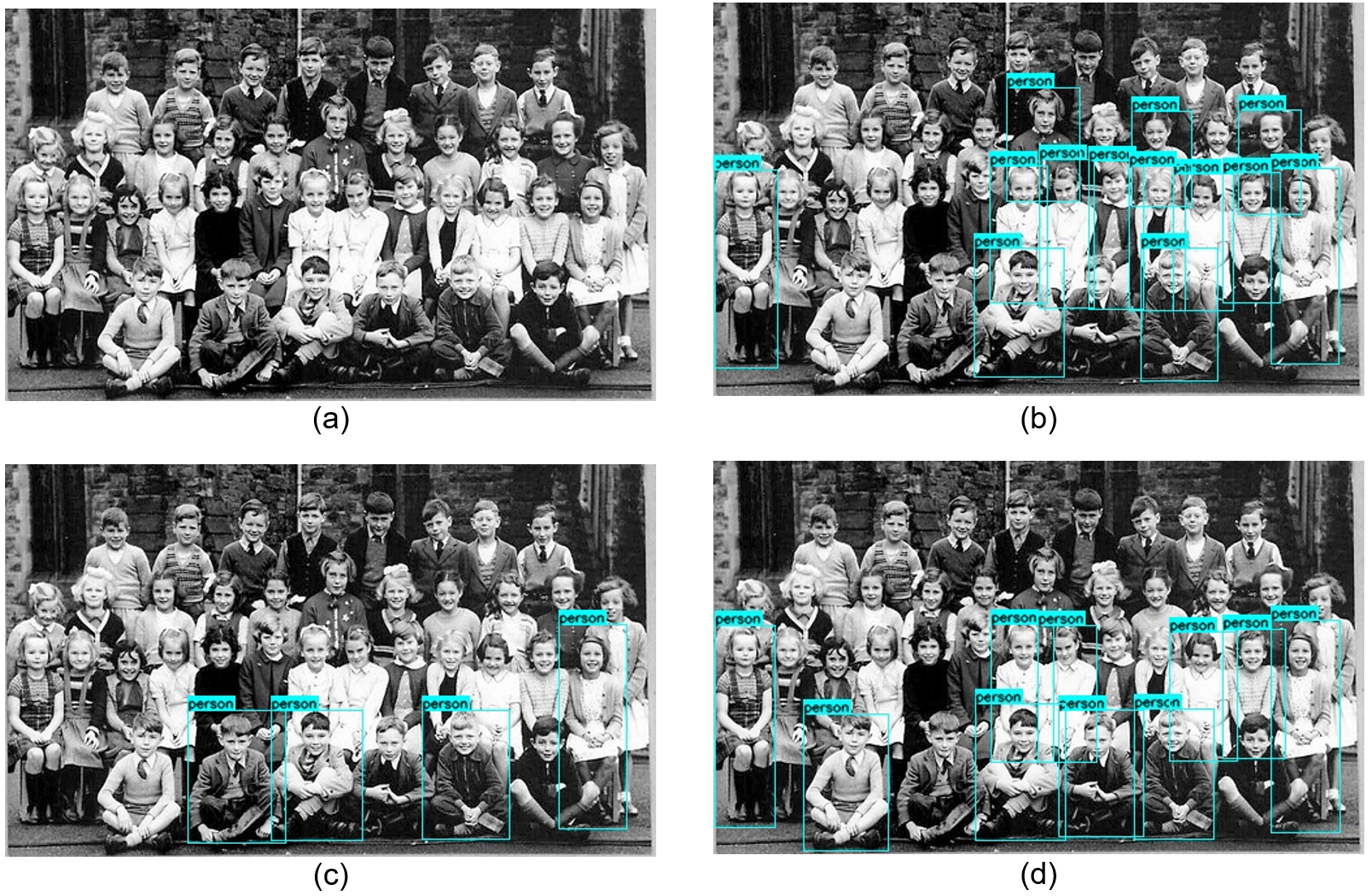}
    \caption{Sample images showing an example where distilled student model detects a higher number of objects compared to the supervised student model trained on the COCO dataset: (a) Original image, (b) Teacher, (c) Student trained supervised with ground truth labels, and (d) Student trained using the proposed distillation approach on COCO (no additional unlabeled data was used). Models are the ones from table \ref{tab:5}.}
    \label{fig:existing_methods_1}
\end{figure*}

\begin{figure*}
    \centering
    \includegraphics[width=1.0\linewidth]{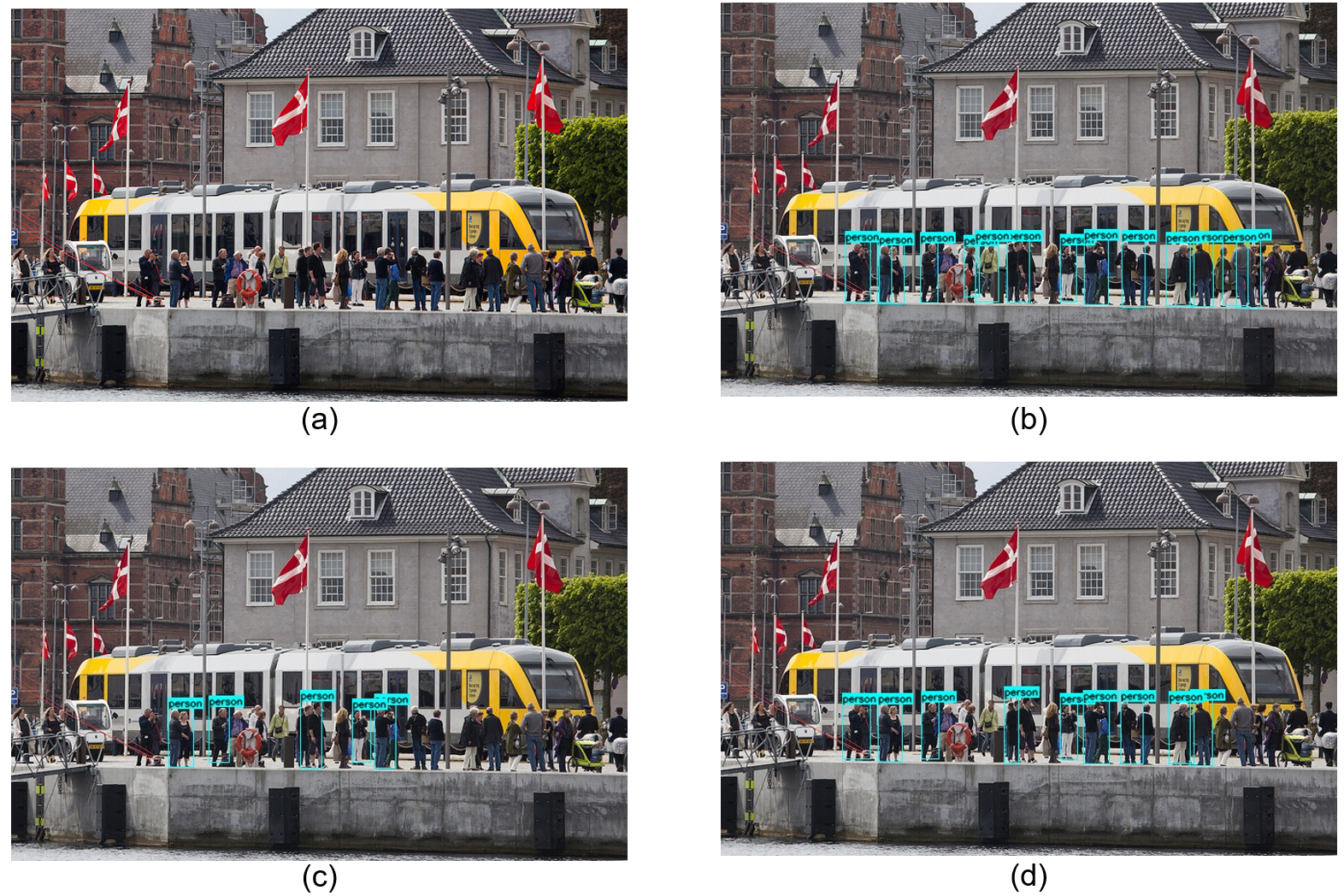}
    \caption{Sample images showing an example where distilled student model detects a higher number of objects compared to the supervised student model trained on the COCO dataset: (a) Original image, (b) Teacher, (c) Student trained supervised with ground truth labels, and (d) Student trained using the proposed distillation approach on COCO (no additional unlabeled data was used). Models are the ones from table \ref{tab:5}.}
    \label{fig:existing_methods_2}
\end{figure*}

Fig. \ref{fig:existing_methods_1} and Fig. \ref{fig:existing_methods_2} illustrate examples were the student models trained using the proposed method show better detection performance than students trained in a standard supervised manner.

\begin{table} [ht]
    \centering
    \caption{Object detection distillation on Pascal-VOC-C dataset (VOC07 clean train set + VOC12 unlabeled corrupt train set, VOC07 corrupt test set).}\vspace{-6pt}
    \setlength{\tabcolsep}{0.01em}
    \renewcommand{\arraystretch}{1.0}%
    \begin{tabular}{|C{2.2cm}|C{2cm}|C{2cm}|}
    \hline
        Model & mAP\\ \hline \hline
        Supervised & 53.78 \%\\ \hline
        Ours & \textbf{62.52} \% \\
    \hline
    \end{tabular}
    \label{tab:sup_2}
\end{table}

\begin{table*}
    \centering
    \caption{Object detection distillation for combining models with different architectures trained on different data, distilled to a new architecture}\vspace{-6pt}
    \setlength{\tabcolsep}{0.01em}
    \renewcommand{\arraystretch}{1.0}%
    \begin{tabular}{|C{5.2cm}|C{6.4cm}|C{3.4cm}|C{2cm}|}
    \hline
        Model & Train Set (size) & Val Set (size) & mAP50:95\\ \hline \hline
        Teacher1: EffDet-D1 & COCO (118K) & COCO (5K) & 38.89 \%\\ \hline
        Teacher2: EffDet-D0 & VOC07+12 (16.5K) & VOC07 test (5K) & 55.22 \% \\ \hline
        Baseline Student: RetinaNet-R50 & COCO + VOC (135K) & COCO + VOC07 (10K) & 34.05 \% \\ \hline \hline
        \textbf{Ours:} UD\textsubscript{COCO+VOC} + FT\textsubscript{COCO+VOC} & COCO + VOC (130K pseudo, 135K labeled) & COCO + VOC07 (10K) & \textbf{37.98}\%\\ \hline
        \textbf{Ours:} UD\textsubscript{OID} + FT\textsubscript{COCO+VOC} & OID (1.4M pseudo), COCO+VOC (135K) & COCO + VOC07 (10K) & \textbf{35.04}\%\\  \hline
    \end{tabular}
    \label{tab:sup_3}
\end{table*}


